\documentclass[letterpaper, 10 pt, conference]{ieeeconf}  

\IEEEoverridecommandlockouts                              

\usepackage{graphics} 
\usepackage{epsfig} 
\usepackage{amsmath} 
\usepackage{amssymb}  
\usepackage{tikz}
\usetikzlibrary{decorations.pathmorphing}  
\usetikzlibrary{decorations.pathreplacing} 
\usepackage{multirow}
\usepackage{subfigure}

\usepackage{lipsum}

\title{\LARGE \bf
CATCH-FORM-ACTer: Compliance-Aware Tactile Control and Hybrid Deformation Regulation-Based  
Action Transformer \\ for  Viscoelastic Object Manipulation}

\author{Hongjun Ma, Weichang Li, Jingwei Zhang, Shenlai He, and Xiaoyan Deng. 
\thanks{*This work is supported by the National Nature Science
Foundation of China under Grant (62473158), Guangdong Basic and Applied Basic Research Foundation under Grant (2022B1515120017, 2023A1515011836).}
\thanks{The authors are all affiliated with with School of Automation Science and Engineering,
        South China University of Technology, 510641, Guangzhou, China; Institute for Super Robotics (Huangpu), 510700, Guangzhou, China. Corresponding author’s email:  {\tt\small mahongjun@scut.edu.cn}}%
}

\begin{document}

\maketitle
\thispagestyle{empty}
\pagestyle{empty}

\begin{abstract}

Automating contact-rich manipulation of viscoelastic objects with rigid robots faces challenges including dynamic parameter mismatches, unstable contact oscillations, and spatiotemporal force-deformation coupling. 
In our prior work, a Compliance-Aware Tactile Control and Hybrid Deformation Regulation (CATCH-FORM-3D) strategy  fulfills robust and effective manipulations of 3D viscoelastic objects, which combines a contact force-driven admittance outer loop and a PDE-stabilized inner loop, achieving sub-millimeter surface deformation accuracy and $\pm5\%$ force tracking.
However, this strategy requires fine-tuning of object-specific parameters and task-specific calibrations, to bridge this gap,  a CATCH-FORM-ACTer is proposed, by enhancing CATCH-FORM-3D with a framework of Action Chunking with Transformer (ACT).
 An intuitive teleoperation system performs Learning from Demonstration (LfD)  to build up a long-horizon  sensing, decision-making and execution sequences. Unlike conventional ACT methods focused solely on trajectory planning, our approach dynamically adjusts stiffness, damping, and diffusion parameters in real time during multi-phase manipulations, effectively imitating human-like force-deformation modulation. Experiments on single arm/bimanual robots in three tasks  show better force fields patterns and thus $10\%-20\%$ higher success rates versus conventional methods, enabling precise, safe interactions for industrial, medical or household scenarios. 



\end{abstract}

\section{INTRODUCTION}

Vision-tactile perception is crucial for dexterous robotic manipulation in unstructured environments \cite{survey}. We focus on rigid robots (e.g., industrial arms and collaborative robots) with position/velocity control, enhanced by compliance schemes integrating visual feedback (e.g., RGB-D cameras) and tactile sensors (e.g., optical tactile skins). While enabling basic contact-rich tasks, it remains challenging to achieve human-like dexterity with viscoelastic materials (e.g., polymers, biomaterials)  due to the
interplay of energy storage, viscous dissipation, and stress redistribution \cite{survey1}-\cite{c1}. Advances in visual-tactile fusion and physics-aware control are needed to predict material behavior and enable precise, adaptive manipulation under real-world variability.


\begin{figure}[ht]
    \centering
    \includegraphics[scale=0.25]{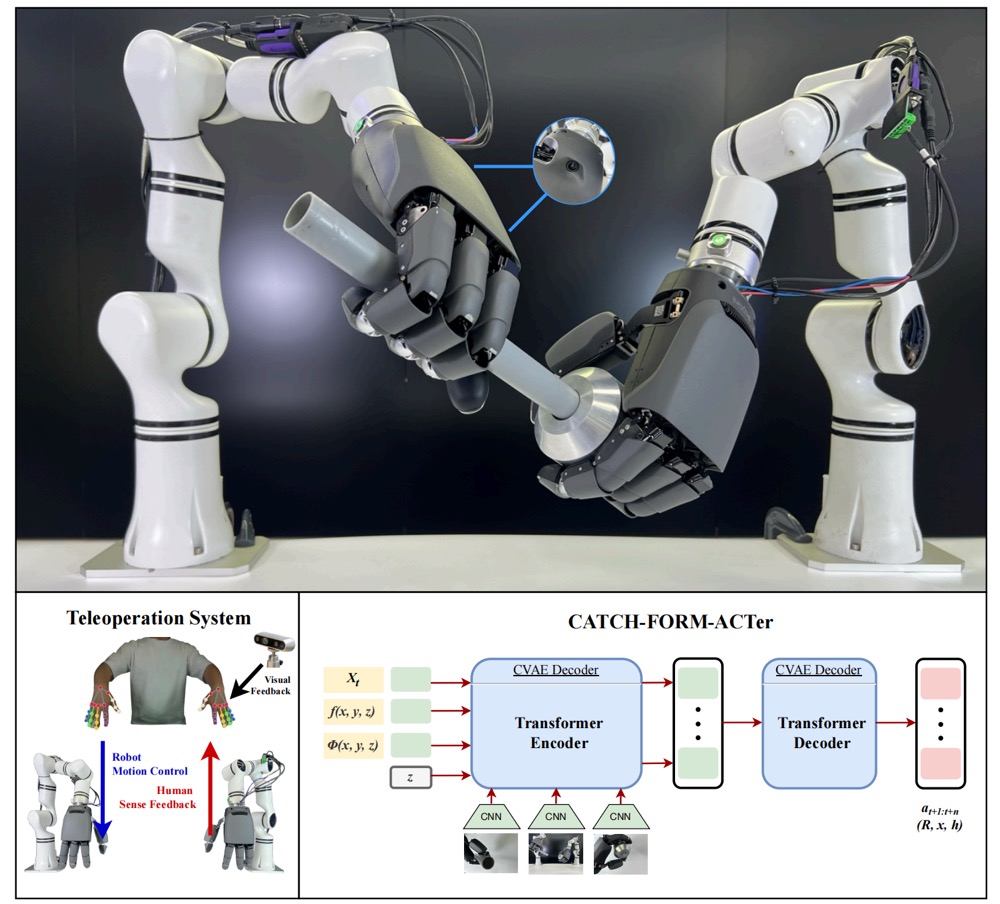}   
    \caption{Our system integrates a teleoperation interface using 3D MoCap controllers, enhanced with visual-cue feedback and safety mechanisms (bottom left), and an LfD framework powered by CATCH-FORM-ACTer (bottom right). The proposed approach was validated on a dual-arm robotic setup,  across complex, contact-rich manipulation tasks (top).}
    \label{fig:controlsystem}
    \vspace{-6mm}
\end{figure}

Contact-rich manipulation demands real-time adaptation to object property variations, making effective algorithms and learning models \cite{cr}-\cite{cr1}, essential for advancing robotic dexterity and autonomy. In our prior work \cite{our}, we proposed a PDE-driven observer to capture spatiotemporal stress-strain dynamics (3D Kelvin–Voigt and Maxwell models) and an inner-outer admittance control architecture for precise force and deformation control. However, this approach overlooked the multi-phase nature of viscoelastic object manipulation (e.g., polymers, biological tissues), which typically includes initial contact, deformation control, steady-state holding, and release. Addressing transitional phases is critical for adaptive manipulation in real-world tasks.
So we integrate a phase-aware modality LfD strategy into our concurrent work, enabling dynamic  compliance adaptation regulating elastic (instantaneous deformation) and viscous (time-dependent relaxation) behaviors across different task phases.


 Robots learn tasks by observing human demonstrations  in LfD, eliminating the need for explicit coding and enabling fine motor control and adaptive force modulation \cite{cr2},\cite{cr3}. This method \cite{surveyc} bridges the gap between high-level task understanding and low-level robotic execution, making robot training more efficient in contact-rich environments.
However, LfD faces challenges such as sample inefficiency, optimal parameter selection, and the need for intuitive teaching interfaces. 
For phase-aware manipulation skill learning, we propose a demonstration system for contact-rich tasks that combines a user-friendly interface that automates parameter tuning, and ensures effective knowledge transfer.
  

Our system (Fig.1) employs a teleoperation interface with 3D motion capture (MoCap) controllers as the primary input device \cite{anytele}, enabling direct robot control for task demonstration. The platform supports cross-robot compatibility, enhancing versatility for diverse manipulation tasks. While based on existing systems \cite{tele1}-\cite{tele3}, our innovation lies in real-time visual cues that alert operators to critical thresholds (joint limits, speed constraints, contact forces and surface deformation limits), ensuring high-quality data collection. A basic compliance control
is integrated further safeguarding against excessive contact forces. This synergy of intuitive MoCap operation, failure-aware feedback, and embedded safety establishes   a versatile and accessible platform.

 
We introduce CATCH-FORM-ACTer, a phase-aware manipulation strategy using Action Chunking with Transformers. The approach collects multi-modal data, including end-effector trajectories, visual-tactile sensor inputs (e.g., surface deformation, contact forces), compliance control parameters (e.g., stiffness, damping, diffusion coefficients), and multi-view images. A transformer-based policy predicts action chunks (short-horizon sensing, decision-making-execution primitives) based on the robot’s current state. Unlike \cite{act}-\cite{act3}, our method emphasizes three innovations: task-space learning for spatiotemporal viscoelasticity, force/deformation conditioning for contact dynamics adaptation, and stiffness-damping-diffusion co-learning to adjust compliance parameters alongside motion trajectories. This framework enables long horizon compliance control and robust handling of complex viscoelastic interactions under dynamic uncertainties.


In summary, we make the following contributions:

1. Effective fundamental controller:  all parameters have clear physical meanings, corresponding to the key mechanical properties of viscoelastic object manipulation, ensuring interpretability and tunability of the control strategy with
robustness and stability in complex contact scenarios.

2. CATCH-FORM-ACTer policy: a transformer-based method integrating ACT with task-space learning and stiffness-damping-diffusion co-adjustment, enabling phase-aware compliance control from minimal demonstrations.

3. Low-cost teleoperation interface: a 3D MoCap system with integrated visual-cues feedback and safety mechanisms  for reliable, intuitive teaching of contact-rich tasks.  

4. Experimental validation: demonstrated on single-arm and bimanual setups across bio-inspired materials (e.g., silicone, polymer foam), achieving better force fields patterns and $10\%-20\%$ higher success rates than existing methods.

 \section{ RELATED WORK}

 \subsection{Viscoelastic Material Manipulation}

 Viscoelastic material manipulation has seen significant advancements in recent years, driven by applications in soft robotics, biomedical engineering, and advanced manufacturing. Researchers have developed precise control strategies \cite{survey} to handle the complex rheological properties of these materials, which exhibit both viscous and elastic behaviors. Techniques such as magnetic actuation, acoustic levitation, and microfluidic systems \cite{survey1} were employed to manipulate viscoelastic substances at various scales. Computational modeling and machine learning \cite{c8}, \cite{c9} enhanced the understanding of material responses under different conditions, enabling more accurate predictions and control. But, challenges remain in achieving real-time feedback and scalability for industrial, physical or household applications, etc..  
 Our prior work \cite{our} proposes a dual-loop control framework for viscoelastic material manipulation, unifying 3D Kelvin–Voigt and Maxwell dynamics in a PDE model. A PDE-driven observer estimates mechanical parameters in real time via visual-tactile fusion. The admittance-based outer loop adapts deformation using force feedback, while the inner loop stabilizes tracking errors via reaction-diffusion PDE boundary control.  This paper is to further address dynamic parameter mismatches, unstable contact oscillations  and spatiotemporal force-deformation couplings, caused by multi-phase practical viscoelastic object manipulation.
 

\vspace{-0.5mm}

 \subsection{ Learning from Demonstrations}

Learning from Demonstrations (LfD) \cite{cr} allows robots to learn skills by imitating humans, with advances enhancing generalizability, adaptability, and efficiency. For example, task-parameterized LfD \cite{cr1}-\cite{cr3} was developed to encode contextual information into reference frames, enhancing skill generalization across different environmental conditions. Recently, new algorithms \cite{act}-\cite{act3} have been proposed to learn from limited demonstrations, leveraging reference frame weights to capture task relevance and improve skill acquisition. In contrast, we propose a LfD framework specially considering that viscoelastic manipulation faces challenges like high sample inefficiency, parameter tuning (e.g., compliance control), and limited teaching interfaces.
Our solution integrates an intuitive interface with automates parameter adaptation (e.g., stiffness/damping/diffusion adjustments), providing
a  robust and effective skill transfer across varying material properties and environmental conditions. Different to existing  ACT frameworks
using trajectory control,  multi-modal data (end-effector trajectories, visual-tactile inputs, compliance parameters, multi-view images) is sampled to train a  transformer, CATCH-FORM-ACTer, predicting force-deformation action chunks  for long horizon manipulations.

\section{CONTROL SYSTEM}

\subsection{Robotic Arm and End-Effector}

Our experimental platform (as Fig.\ref{fig:controlsystem}) comprises two Realman RM65-B arms, each integrated with built-in force-torque sensors to enable precise force measurement during manipulation tasks. The system employs PaXini DexH13 (13 jointed dexterous hand) as end-effectors, which are equipped with high-resolution visual-tactile sensors embedded in both the fingertips and palms. These sensors feature a 12$\times$10 tactile element matrix capable of capturing localized pressure distributions (0–50kPa) and surface displacements at a sampling rate of 200Hz, delivering rich real-time tactile feedback for dynamic object interactions. Force fields are calculated through directed surface projections, while deformation fields are generated by applying Poisson filters to point cloud data, enabling detailed characterization of contact mechanics.

The visual perception is integrated by a multi-camera setup, that is, two RGB-D cameras are embedded in the palms of the PaXini hands for close-range, task-centric observation, and a third static RGB-D camera monitors the entire workspace. This hierarchical configuration ensures comprehensive visual coverage across multiple perspectives, enhancing state estimation accuracy during complex manipulation sequences. The palm-mounted cameras prioritize fine-grained object and contact monitoring, while the global camera supports spatial awareness and trajectory planning.

\subsection{Teleoperation Interface}
 
 \begin{figure}[h]
    \centering
    \includegraphics[scale=0.12]{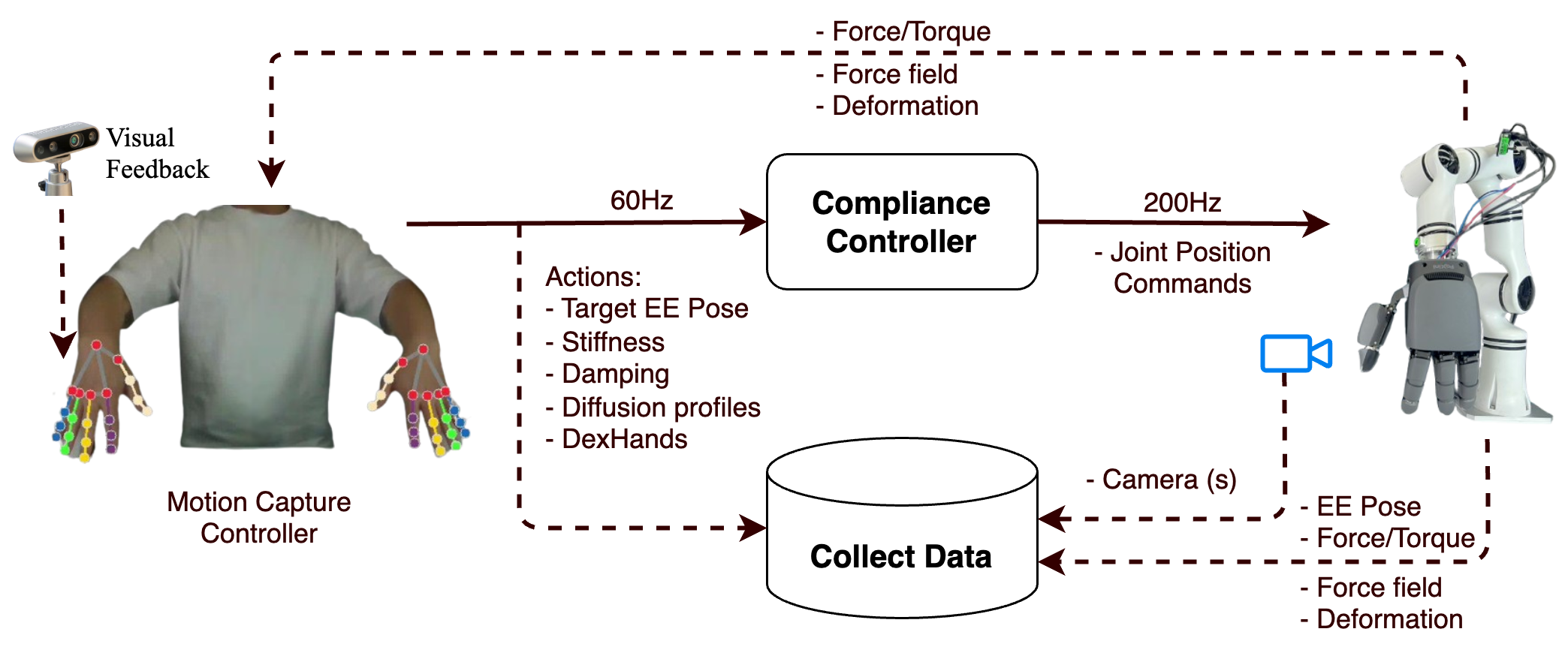}   
    \caption{ Teleoperation system for data collection using 3D MoCap
controllers. The controller receives the contact force and surface deformation
information from the robot’s finger sensors and uses it to
provide visual cues feedback to the operator. The same concept is applied to bimanual tasks
where a different 3D MoCap controller is connected to each arm.}
    \label{fig:framework}
\end{figure}

Our teleoperation system (Fig.\ref{fig:framework}) employs 3D motion capture (MoCap) controllers inspired by the AnyTeleop framework \cite{anytele} to enable intuitive dexterous hand-arm manipulation, operating through a client-server pipeline that processes RGB-D camera streams at 60Hz for markerless hand pose detection. The detection module outputs two critical data streams: local finger keypoint positions within the wrist frame, capturing fine-grained finger articulations (i.e., two hands' 26 joint angles for precision grasping), and global 6D wrist pose (position and orientation) in the camera frame, providing spatial context for arm movement. 
These data is converted into joint commands using a real-time inverse kinematics solver optimized for dexterous manipulators, with commands transmitted via UDP under a low-latency protocol ($<$10ms delay) to synchronize robot motion. Relative hand motion (calculated from a keyboard-defined reference pose) is dynamically scaled using adaptive motion mapping algorithms that account for workspace disparities (e.g., human-to-robot arm length ratios of 1:1.5), ensuring intuitive control despite kinematic differences. 
Real-time metrics like Max. contact force and Max. deformation, are displayed alongside a dynamic force-deformation visualization, providing operators with actionable insights. The  controller leverages these fields data and stiffness-damping-diffusion parameters to perform a compliance with default values for the parameters, based on which operators are able to make further fine adjustments during demonstrations, particularly for tasks requiring precise force-deformation modulation.

 \subsection{Compliance Control via CATCH-FORM-3D}

Our CATCH-FORM-3D policy proposed a physics-guided admittance control framework that explicitly addresses the regulation of reference deformations during interactions with viscoelastic objects. The framework integrates a unified 3D viscoelastic continuum model combining Kelvin–Voigt and Maxwell dynamics to govern energy storage, viscous dissipation, and stress redistribution across multi-parameter fields. An observer-based parameter identification method dynamically estimates material properties by regressing historical actuation forces and visual-tactile signals, while a physics-guided planning strategy interplays deformation modulation with compliant force regulation through viscoelastic dynamics. To ensure stability under large deformations, a boundary control strategy synthesizes Dirichlet boundary conditions via analytical geometric templates, guaranteeing globally convergent strain fields and force stabilization. An inner-outer admittance control architecture achieves low-error force profiles and precise geometry transformations (sub-millimeter accuracy) in dynamic tasks, balancing computational efficiency (10ms cycle time) with physical fidelity. 
The controller parameters include $\epsilon_1$, $\lambda_1$, $\lambda_2$ which are used to tune the stiffness, damping and diffusion
of the closed-loop system (more detailed illustration is available in \cite{our}).

\section{CATCH-FORM-ACTer: LEARNING PHASE-\\AWARE COMPLIANCE
CONTROL FROM A FEW DEMONSTRATIONS}

 \begin{figure*}[htbp]
    \centering
    \includegraphics[width=\linewidth]{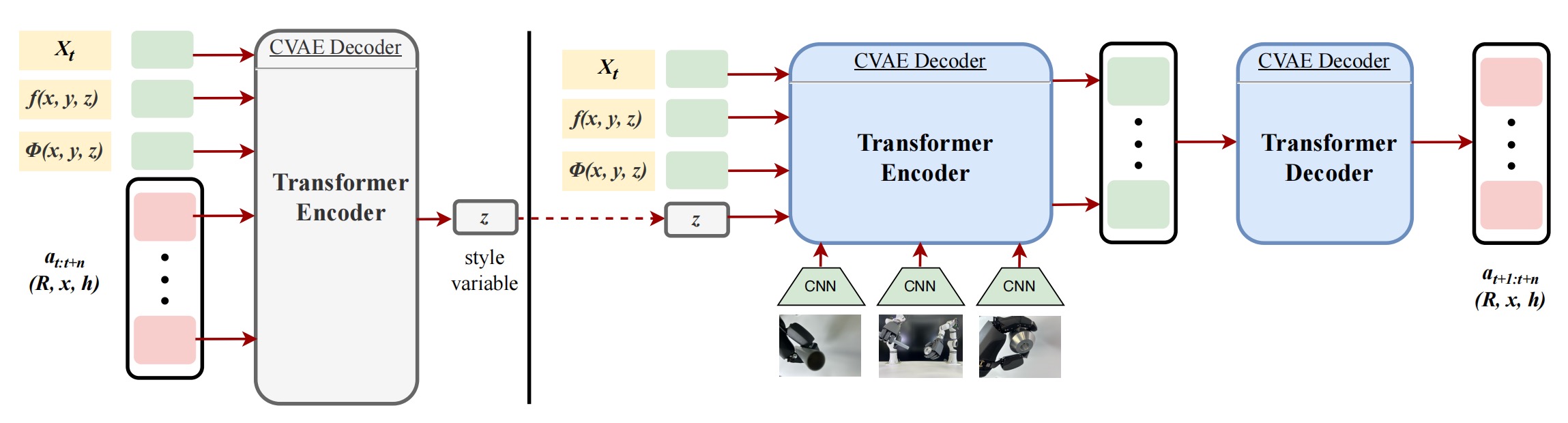}   
    \caption{ CATCH-FORM-ACTer Network architecture. The action sequence, consisting of n robot states (stiffness/damping/diffusion parameters $R$, target EE's Cartesian pose $x$, and dexterous hand joint angles $h$), are encoded alongside the current Cartesian pose $X_t$, contact force field $f(x,y,z)$, and surface deformation field $\phi(x,y,z)$ by the CVAE encoder. This
network is discarded at inference time. Right: The policy inputs are images from multiple viewpoints, the current Cartesian pose, and the measured force and deformation
fields. The policy predicts a sequence of $n$ future actions.}
    \label{fig:transformer}
\end{figure*}

The CATCH-FORM-ACTer system (Fig.\ref{fig:transformer}) enables phase-aware compliance control for robotic manipulation by learning adaptive strategies from limited demonstrations. Utilizing an Action-Conditioned Transformer (ACT), the system plans multi-dimensional action sequences that simultaneously predict: target end-effector 6D poses and dynamically adjusted compliance parameters (stiffness, damping, diffusion coefficients) based on real-time sensory inputs. Its dual-output architecture allows synchronized motion trajectory planning and physical interaction optimization, achieving context-aware stiffness-damping-diffusion modulation (e.g., reducing rigidity during delicate contact or enhancing precision in positioning) while preserving task-specific kinematic constraints.
By embedding compliance dynamics directly into the action prediction loop, CATCH-FORM-ACTer integrates high-level task reasoning with real-time physical responsiveness, essential for handling viscoelastic materials or unstructured environments. The predicted parameters drive a compliance controller, enabling autonomous adaptation to different task phases (e.g., contact initiation, force modulation, object release) without compromising motion tracking accuracy. This approach eliminates the need for exhaustive demonstrations or manual parameter tuning, effectively bridging learning-from-demonstration (LfD) frameworks with physics-grounded compliance control.

 \subsection{ Transformer-Based Imitation Learning}


Based on the Action Chunking with Transformers (ACT) framework \cite{act}, our implementation extends its capability to jointly learn end-effector (EE) trajectories, and time-varying compliance parameters (stiffness, damping, diffusion coefficients) from sparse demonstrations. The policy is structured as the decoder of a conditional variational autoencoder (CVAE), where the encoder compresses high-dimensional observations, including RGB images, force/deformation tactile sensor arrays (12$\times$10 spatial resolution), and proprioceptive joint states, into a latent distribution. The decoder then autoregressively generates action chunks conditioned on this latent space, with each chunk comprising:  1) 6D wrist pose: Expressed as a 3D positional vector and 3D axis-angle orientation (normalized to $[-\pi, \pi]$ for continuity); 2) Finger articulation: A 13-dimensional vector specifying target joint angles for the PaXini hand’s adaptive grasp synergy control; 3) Compliance parameters: A 3D vector defining stiffness $\lambda_1 \in [50, 500]~N/m$, damping $\lambda_2 \in [0.1, 5.0]~Ns/m$, and diffusion $\epsilon \in [0.01, 0.1]~m^2/s$ coefficients.   This results in a 22-dimensional action space for each robotic arm, with bimanual coordination raising the total dimensionality to 44. 
As illustrated in Fig.\ref{fig:transformer}, the CVAE’s training objective combines a reconstruction loss (minimizing $L_2$ error between predicted and demonstrated actions) with a KL-divergence term regularizing the latent space. Crucially, the compliance parameters are dynamically scaled using sigmoid activations to enforce physical constraints, while axis-angle orientations are converted to rotation matrices via Rodrigues’ formula for stable gradient propagation. The predicted actions are executed by the CATCH-FORM-3D controller.
By chunking actions into 0.1s intervals (10Hz frequency), the system balances temporal abstraction with real-time reactivity, enabling seamless transitions between task phases such as contact exploration (low stiffness, high diffusion) and precision placement (high stiffness, low damping). This design eliminates the need for manual phase detection or parameter tuning, as the policy intrinsically learns phase-dependent compliance strategies through the CVAE’s latent task embedding.

\section{EXPERIMENTAL VALIDATION}

\begin{figure*}[htbp]
\centering
\begin{subfigure}{}
    \includegraphics[width=41mm,clip]{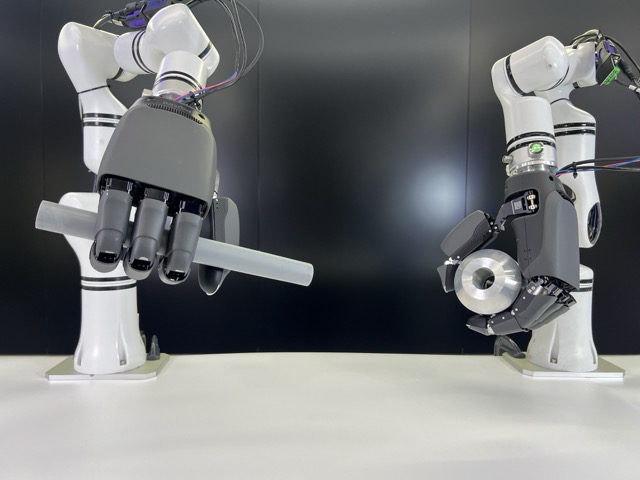}
    \label{figure:task_1a}
\end{subfigure}
\hfill
\begin{subfigure}{}
    \includegraphics[width=41mm,clip]{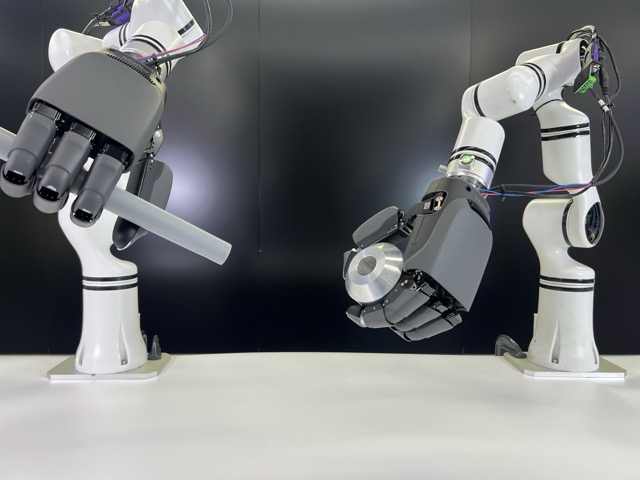}
    \label{figure:task_1b}
\end{subfigure}
\hfill
\begin{subfigure}{}
    \includegraphics[width=41mm,clip]{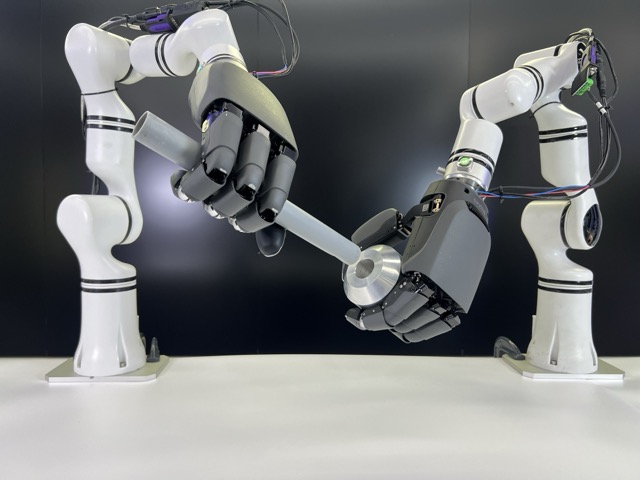}
    \label{figure:task_1c}
\end{subfigure}
\hfill
\begin{subfigure}{}
    \includegraphics[width=41mm,clip]{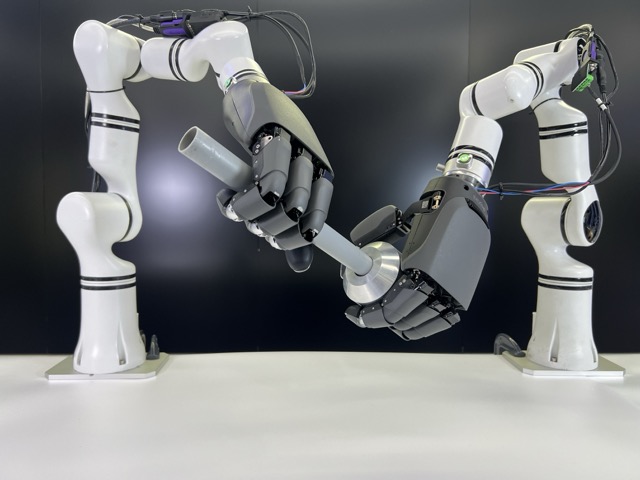}
    \label{figure:task_1d}
\end{subfigure}
\caption{\textbf{Bimanual Insertion (Cylinder)}: This task involves inserting a 3D-printed peg into a matching hole with a 2 mm tolerance. Precise alignment of both arms in position and orientation is required to ensure successful insertion. During motion without contact, both arms operate in medium compliance parameters mode. As insertion begins, the left arm switches to high compliance parameters to maintain stability, while the right arm transitions to low compliance parameters, allowing force guidance to assist the insertion process. A total of 20 demonstrations were performed for this task.}
\label{figure:task_1}
\end{figure*}

The CATCH-FORM-ACTer policy was evaluated on three challenging contact-rich manipulation tasks, as shown in Fig.\ref{figure:task_1} to Fig.\ref{figure:task_3} provide detailed descriptions of these tasks. A summary of the experimental conditions for each task is presented in Table~\ref{table:success_rate}.
For each task, the policy was trained from scratch over 80,000 epochs, which took between 6 to 10 hours of real time, depending on factors such as the number of demonstrations and cameras used.
Unlike previous methods that rely solely on scalar force/torque measurements, our approach leverages rich spatiotemporal information from both contact force fields $f(x,y,z)$ and surface deformation fields $\Phi(x,y,z)$, as illustrated in Fig.\ref{figure:task_3}. As shown in Fig.\ref{figure:task_3}(left), our system employs a four-fingered robotic hand equipped with high-resolution tactile sensors to capture detailed force distributions during manipulation.

\begin{figure*}[htbp]
\centering
\begin{subfigure}{}
    \includegraphics[width=41mm,clip]{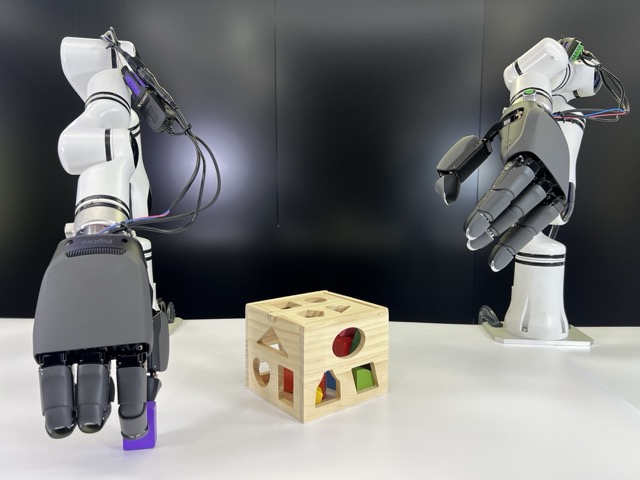}
    \label{figure:task_2a}
\end{subfigure}
\hfill
\begin{subfigure}{}
    \includegraphics[width=41mm,clip]{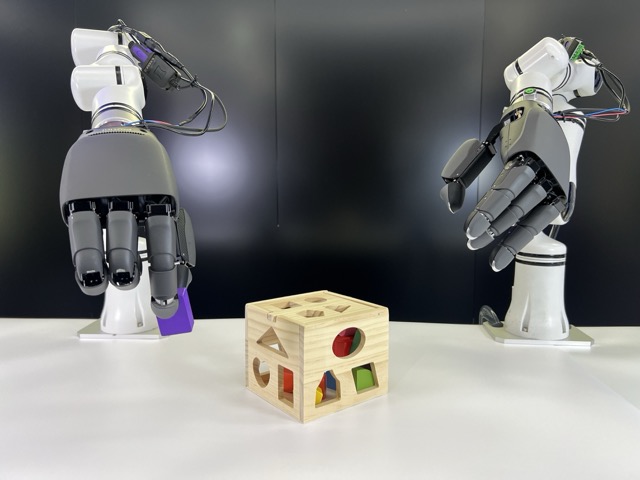}
    \label{figure:task_2b}
\end{subfigure}
\hfill
\begin{subfigure}{}
    \includegraphics[width=41mm,clip]{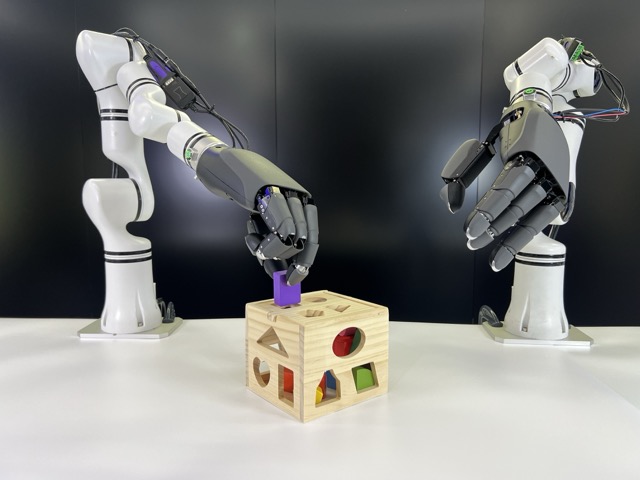}
    \label{figure:task_2c}
\end{subfigure}
\hfill
\begin{subfigure}{}
    \includegraphics[width=41mm,clip]{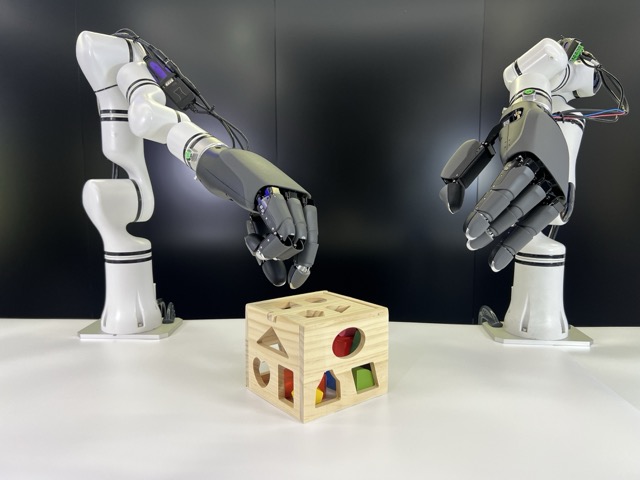}
    \label{figure:task_2d}
\end{subfigure}
\caption{\textbf{Single-Arm Picking $\&$ Insertion}: The task involves picking up a toy wooden peg and inserting it into a corresponding hole in a wooden box. A stable grasp is necessary to align the peg properly with the hole. During demonstrations, the robot operates in medium compliance parameters mode for general manipulation and switches to low compliance parameters mode during the insertion phase, where physical contact occurs. The peg is placed at the edge of the whiteboard with a random rotation of approximately $\pm15^\circ$. The wooden box's position is not fixed, allowing the robot to move it during insertion. A total of 30 demonstrations were conducted for this task.}
\label{figure:task_2}
\end{figure*}

\begin{figure*}[htbp]
\centering
\begin{subfigure}{}
    \includegraphics[width=41mm,clip]{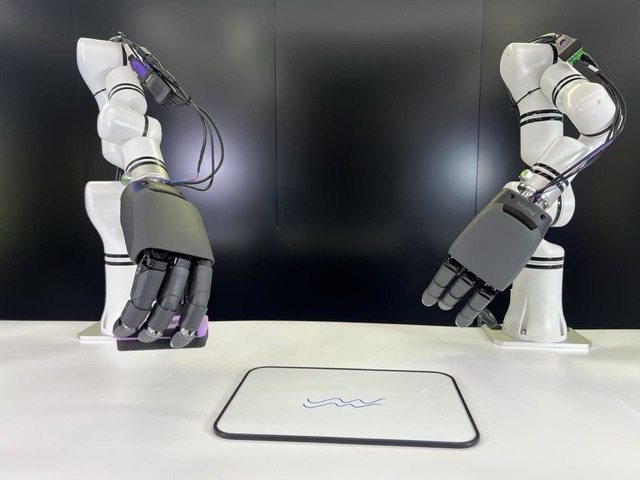}
    \label{figure:task_3a}
\end{subfigure}
\hfill
\begin{subfigure}{}
    \includegraphics[width=41mm,clip]{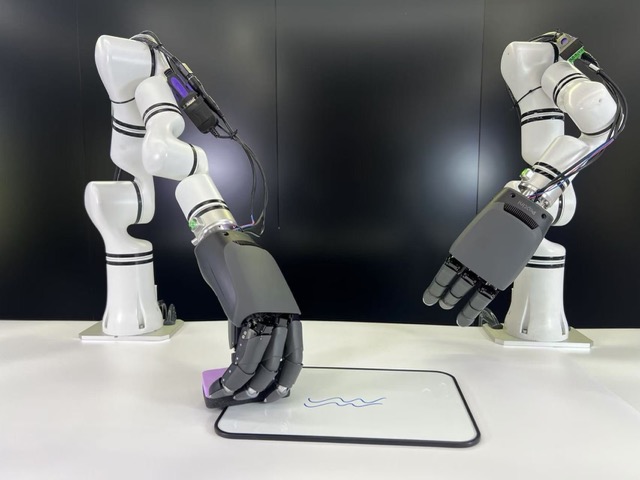}
    \label{figure:task_3b}
\end{subfigure}
\hfill
\begin{subfigure}{}
    \includegraphics[width=41mm,clip]{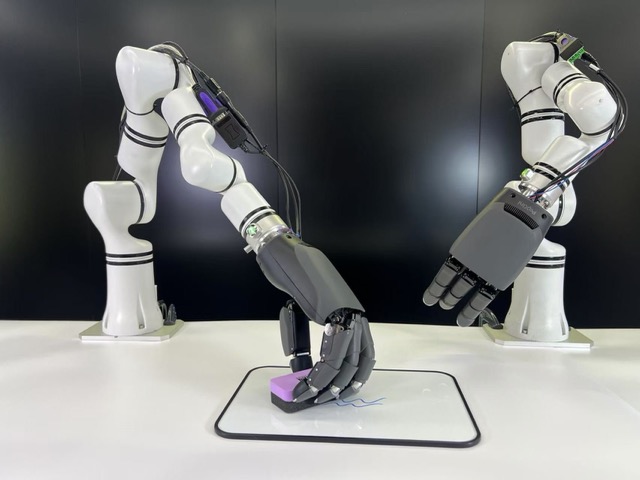}
    \label{figure:task_3c}
\end{subfigure}
\hfill
\begin{subfigure}{}
    \includegraphics[width=41mm,clip]{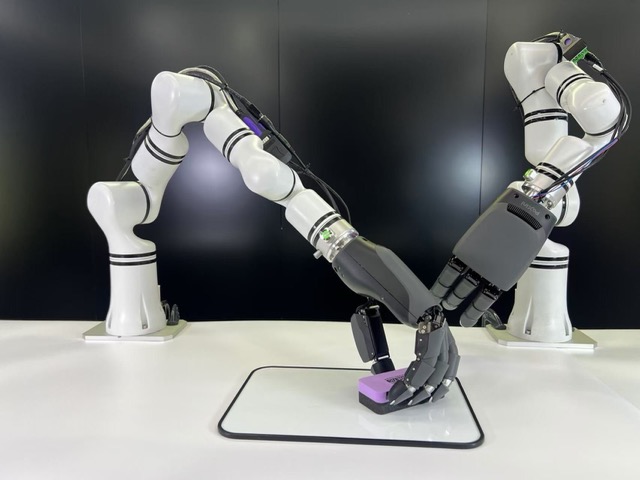}
    \label{figure:task_3d}
\end{subfigure}
\caption{\textbf{Single-Arm Wiping}: The goal of this task is to wipe marks off a whiteboard. The robot starts with the eraser already in hand. During demonstrations, it operates in medium compliance parameters mode while approaching the board, then switches to low compliance parameters mode to apply the necessary force for wiping. Sufficient contact force is required to ensure effective cleaning during the wiping motion. Each trial begins with a randomly placed mark within $\pm5$ cm of the board's center. A total of 20 demonstrations were conducted for this task.}
\label{figure:task_3}
\end{figure*}

\begin{figure*}[htbp]
\centering
\begin{subfigure}{}
    \includegraphics[width=41mm,height=40mm]{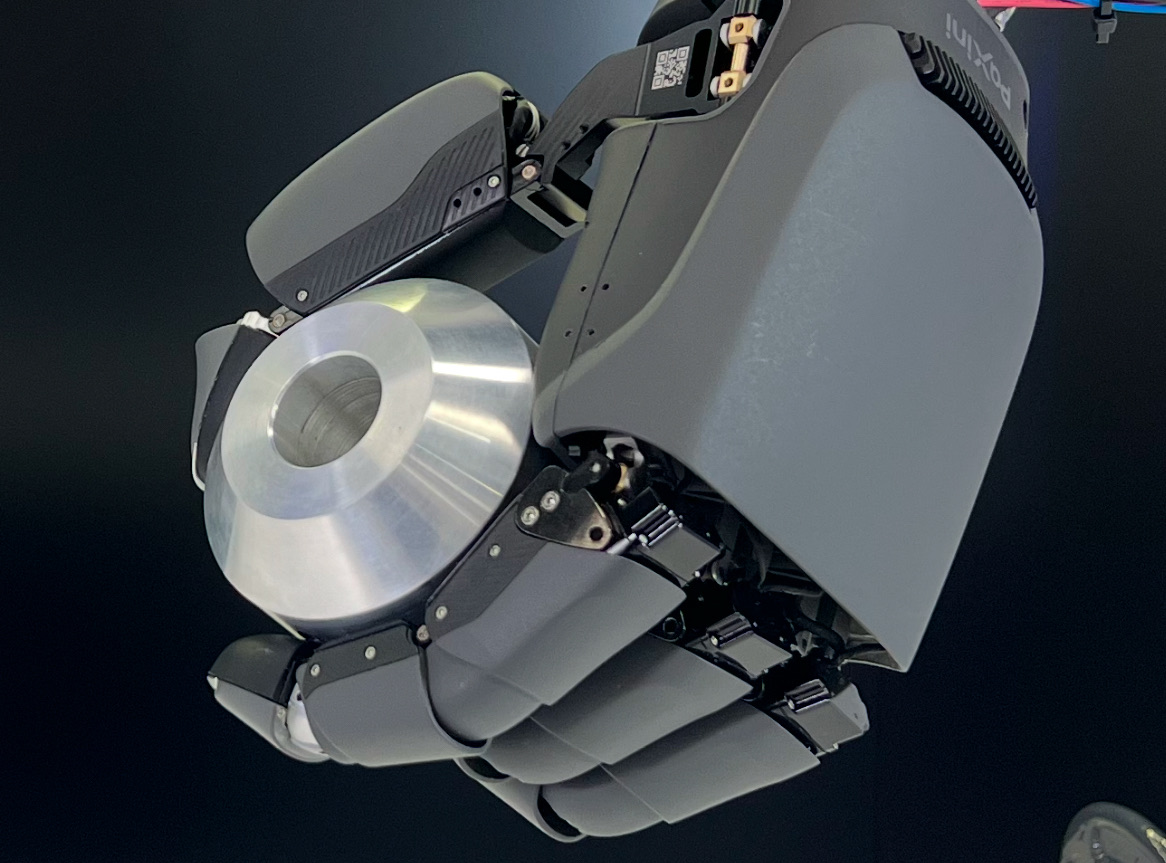}
    \label{figure:Physical gripper setup}
\end{subfigure}
\hfill
\begin{subfigure}{}
    \includegraphics[width=41mm,clip]{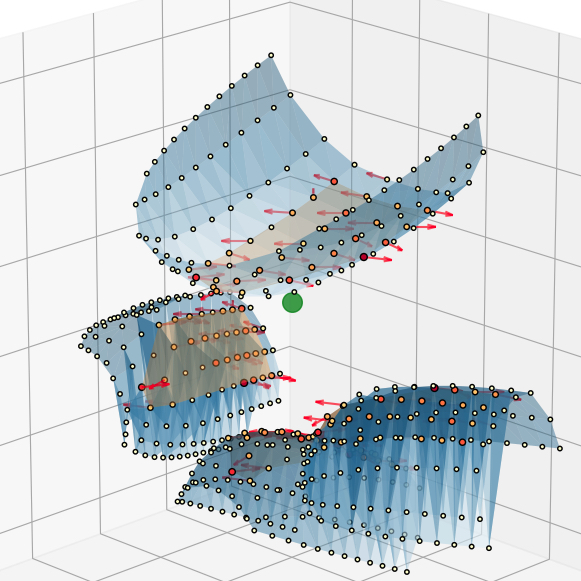}
    \label{figure:force_field_viz_success}
\end{subfigure}
\hfill
\begin{subfigure}{}
    \includegraphics[width=41mm,clip]{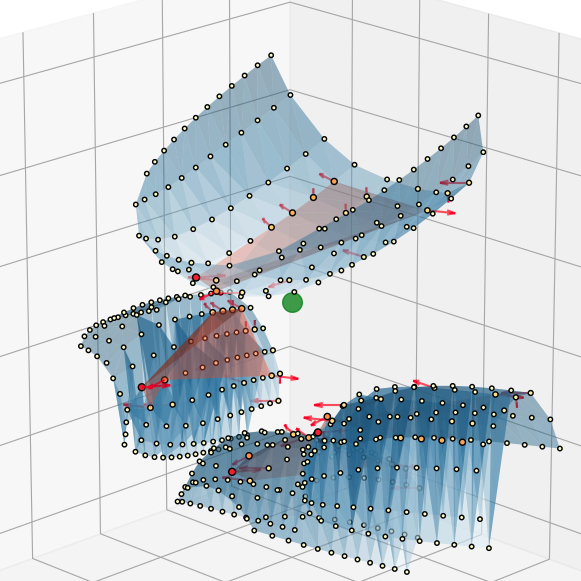}
    \label{figure:force_field_viz_failure}
\end{subfigure}
\caption{Visualization of force fields $f(x,y,z)$ during the bimanual cylinder insertion task (left) (Task 1). Colored surfaces represent the deformation field $\Phi(x,y,z)$ across the four-fingered robotic hand, while red arrows indicate force vectors, during successful insertions (middle) and unsuccessful cases(right).}
\label{figure:task_3}
\end{figure*}


In Task 1 (bimanual cylinder insertion), as depicted in Fig.\ref{figure:task_3} (left), the right hand precisely grasps a 3D-printed peg while the left hand holds the metal target fixture with a circular hole. The visualization in Fig.\ref{figure:task_3} (middle) demonstrates that during successful insertions, our method captures uniform concentric force patterns with balanced pressure distribution across all fingertips of the left hand as it maintains the metal fixture with higher compliance parameters. This enables precise alignment control while the right hand manipulates the peg with lower compliance parameters. In contrast, Fig.\ref{figure:task_3}(right) illustrates how unsuccessful insertions by  ACT and Comp-ACT policy show asymmetric force distribution with imbalanced pressure concentrations on the left hand's fingertips, indicating misalignment that scalar force measurements alone fail to detect. These spatially rich field representations are directly integrated into our transformer-based architecture, allowing the CATCH-FORM-ACTer policy to establish correlations between spatial force/deformation patterns and appropriate compliance parameter adjustments across different manipulation phases.

The tasks we conducted are summarized in the table and include bimanual insertion, single-arm picking and insertion, and single-arm wiping. In the bimanual insertion task, the process can be divided into several phases, including grasping, contact, and insertion. Our method achieved a higher success rate than Comp-ACT in both the grasping and contact phases, demonstrating its effectiveness in handling contact-rich problems. Similarly, in the single-arm picking and insertion and single-arm wiping tasks, we observed that our approach outperforms Comp-ACT in addressing contact-related challenging manipulatioins. It exhibits strong generalization capability for object interactions in real-world scenarios and effectively learns from data. This results in an improved success rate across all three tasks, highlighting the robustness and effectiveness of our method.

\setlength{\tabcolsep}{1.83pt}
\begin{table}[t]
\centering
\caption{Task-Specific Success Rates with Compliance Strategies}
\label{table:success_rate}
\begin{tabular}{|l|c|c|c|c|}
\hline
\multirow{2}{*}{\textbf{Method}} & \multicolumn{3}{c|}{\textbf{Success Rate (\%)}} & \multirow{2}{*}{\begin{tabular}{c}\textbf{Applied} \\ \textbf{Variable} \\ \textbf{Compliance}\end{tabular}} \\
\cline{2-4}
 & \begin{tabular}{c}\textbf{Pick\&Insert} \\ \textbf{(Mid$\rightarrow$Low)}\end{tabular} & \begin{tabular}{c}\textbf{Wiping} \\ \textbf{(Mid$\rightarrow$Low)}\end{tabular} & \begin{tabular}{c}\textbf{Bimanual} \\ \textbf{(L: Mid$\rightarrow$High,} \\ \textbf{R: Mid$\rightarrow$Low)}\end{tabular} & \\
\hline
ACT & 40 & 50 & 40 & - \\
\hline
Comp-ACT & 65 & 70 & 70 & \checkmark \\
\hline
Ours & 85 & 90 & 80 & \checkmark \\
\hline
Ours* & 65 & 75 & 45 & - \\
\hline
\end{tabular}\\
\raggedright\small{*Without force and deformation fields representation}
\end{table}

\subsubsection{Comparative Performance Analysis}

Table~\ref{table:success_rate} presents a comprehensive comparison of our method against ACT and Comp-ACT across all tasks. We observe consistent performance improvements across all real-world scenarios, with especially significant gains in contact-rich tasks.


\textbf{Single-Arm Picking \& Insertion:} For this task (Fig.\ref{figure:task_2}), CATCH-FORM-ACTer achieved an 85\% success rate compared to Comp-ACT's 65\% and ACT's 40\%. The most significant improvement was observed during the insertion phase, where our spatial force field representation allowed the system to detect misalignments much earlier than methods using only scalar force values.

Analysis of the deformation fields during successful insertions revealed that our system learned to correlate specific deformation patterns with optimal insertion trajectories. The policy learned to maintain more uniform deformation patterns across the contact surface by continuously adjusting compliance parameters in response to evolving contact conditions. This resulted in smoother insertions with fewer jamming incidents and reduced peak forces.

\textbf{Single-Arm Wiping:} In this task (Fig.\ref{figure:task_3}), our method achieved a 90\% success rate using full force and deformation field information. Interestingly, the performance improved to 80\% when explicit deformation field information was emphasized in training. This counterintuitive result occurred because the deformation field provided better indicators of effective wiping contact than force information alone.

The wiping task highlights another key advantage of our spatial field approach: during wiping motions, the distribution of forces across the contact surface is more important than the absolute magnitude. The deformation field allowed the system to maintain consistent contact across the entire eraser surface, something that scalar force measurements alone couldn't facilitate. Analysis of unsuccessful trials by Comp-ACT showed that despite applying appropriate overall force, the pressure distribution was often concentrated at the edges of the eraser, resulting in incomplete wiping.

\textbf{Bimanual Insertion (Cylinder):} This task (Fig.\ref{figure:task_1}) represents one of the most challenging operations in our evaluation suite, requiring precise coordination between arms while managing complex contact dynamics. Our method achieved a 80\% success rate when utilizing the full force and deformation field information, compared to 70\% for Comp-ACT and only 40\% for standard ACT.

The key advantage of our approach in this task stems from the system's ability to detect early signs of misalignment through spatial force field analysis. Fig.\ref{figure:task_3} illustrates the difference in force field patterns between successful and unsuccessful insertions. In successful trials, our system detected asymmetric force distributions at initial contact and adjusted the compliance parameters accordingly before excessive forces developed. These adjustments were made specifically in regions of high force concentration, demonstrating the spatial selectivity enabled by our approach.

The results demonstrate that our phase-aware CATCH-FORM-ACTer approach provides significant advantages for contact-rich manipulation tasks, particularly those involving precision alignment, force modulation, and coordination between multiple arms. By learning correlations between spatial force/deformation patterns and appropriate compliance adjustments, CATCH-FORM-ACTer achieves more robust and adaptive behavior than methods relying solely on scalar force measurements or fixed compliance parameters.

\section{CONCLUSIONS}

We studied the problem of learning contact-rich manipulation tasks from a few demonstrations for 
multi-phase manipulation of viscoelastic objects. To this end, we presented a novel teleoperation
system with 3D MoCap-based remote operation and an imitation learning method for learning 
phase-aware  compliance control called CATCH-FORM-ACTer. 
Our proposed system enables users to teleoperate a single arm or bimanual robotic system to safely
collect demonstrations of challenging manipulation tasks.
Our evaluation showed that CATCH-FORM-ACTer could learn complex
tasks with a high success rate from 20 to 30 demonstrations,
which can reduce the risk of excessive
forces by learning the stiffness-damping-diffusion parameters of the underlying
compliance controller, enabling safe task execution with
variable compliance in complex force-deformation interactions.

\addtolength{\textheight}{-12cm}   





\section*{ACKNOWLEDGMENT}

Authors are especially grateful to PaXini Technology for
assisting in the construction of the experimental platform.


\end{document}